\documentclass[10pt,twocolumn,letterpaper]{article}
\date{December 7, 2025}
\usepackage[margin=0.75in,columnsep=0.25in]{geometry}

\usepackage{times}
\usepackage[utf8]{inputenc}
\usepackage[T1]{fontenc}

\usepackage{hyperref}
\usepackage{url}
\usepackage{booktabs}
\usepackage{amsfonts}
\usepackage{amsmath}
\usepackage{amssymb}
\usepackage{nicefrac}
\usepackage{microtype}
\usepackage{xcolor}
\usepackage{graphicx}
\usepackage{algorithm}
\usepackage{algorithmic}
\usepackage{subcaption}
\usepackage{titlesec}
\usepackage[numbers]{natbib}
\usepackage{caption}

\titleformat{\section}{\normalfont\large\bfseries}{\thesection}{1em}{}
\titleformat{\subsection}{\normalfont\normalsize\bfseries}{\thesubsection}{1em}{}
\titleformat{\subsubsection}{\normalfont\normalsize\itshape}{\thesubsubsection}{1em}{}
\titlespacing*{\section}{0pt}{2ex plus 1ex minus .2ex}{1ex plus .2ex}
\titlespacing*{\subsection}{0pt}{1.5ex plus 1ex minus .2ex}{0.5ex plus .2ex}
\titlespacing*{\subsubsection}{0pt}{1ex plus 1ex minus .2ex}{0.5ex plus .2ex}

\usepackage{xurl}

\graphicspath{{media/}}

\title{Task adaptation of Vision-Language-Action model: \\
1st Place Solution for the 2025 BEHAVIOR Challenge}

\author{%
  \begin{tabular}[t]{@{}c@{}}
    Ilia Larchenko\textsuperscript{*} \\
    \texttt{\small ilya.larchenko@gmail.com}
  \end{tabular}
  \hspace{1.5em}
  \begin{tabular}[t]{@{}c@{}}
    Gleb Zarin\textsuperscript{*}\\
    \texttt{\small zaringleb@gmail.com}
  \end{tabular}
  \hspace{1.5em}
  \begin{tabular}[t]{@{}c@{}}
    Akash Karnatak\textsuperscript{*}\\
    \texttt{\small akashkarnatak.pro@gmail.com}
  \end{tabular}
  \vspace{5.0mm}
  \\[1.0ex]
  \textsuperscript{*}Independent Researchers
}

\begin{document}

\twocolumn[
  \begin{@twocolumnfalse}
    \maketitle
    \begin{abstract}
    We present a vision-action policy that won 1st place in the 2025 BEHAVIOR Challenge---a large-scale benchmark featuring 50 diverse long-horizon household tasks in photo-realistic simulation, requiring bimanual manipulation, navigation, and context-aware decision making.

    Building on the Pi0.5 architecture, we introduce several innovations. Our primary contribution is \textbf{correlated noise for flow matching}, which improves training efficiency and enables \textbf{correlation-aware inpainting} for smooth action sequences. We also apply \textbf{learnable mixed-layer attention} and \textbf{System 2 stage tracking} for ambiguity resolution. Training employs \textbf{multi-sample flow matching} to reduce variance, while inference uses \textbf{action compression} and challenge-specific \textbf{correction rules}.

    Our approach achieves 26\% q-score across all 50 tasks on both public and private leaderboards.
    \end{abstract}
    \vspace{1em}
  \end{@twocolumnfalse}
]

\section{Introduction}

\subsection{The BEHAVIOR Challenge}

The BEHAVIOR Challenge presents a demanding benchmark for embodied AI: performing 50 diverse, long-horizon household tasks in photo-realistic simulation. Tasks range from simple (turning on radio) to complex multi-step activities (cooking a hotdog). The challenge requires:

\begin{itemize}
    \item \textbf{Long-horizon execution}: 6.6 minutes on average per task, with the longest tasks averaging 14 minutes
    \item \textbf{Bimanual manipulation}: Coordinated use of two 7-DOF arms with parallel-jaw grippers
    \item \textbf{Mobile navigation}: Indoor navigation through cluttered environments
    \item \textbf{Multi-camera perception}: Processing RGB images from head and both wrist cameras
    \item \textbf{Task diversity}: 50 different activities evaluated with a single policy (or small set of checkpoints)
\end{itemize}

The benchmark uses OmniGibson simulation built on NVIDIA Isaac Sim, providing realistic physics and rendering. Each task is evaluated over 10 episodes with randomized initial conditions, and performance is measured by a q-score that combines success rate with partial credit for subtask completion.

\subsection{Key Challenges}

Long-horizon household manipulation poses several fundamental challenges:

\begin{itemize}
    \item \textbf{Compounding errors.} With episodes spanning thousands of timesteps, small prediction errors can accumulate. This demands either extremely accurate predictions or robust recovery behaviors.
    
    \item \textbf{Non-Markovian states.} Many task states are visually ambiguous---the robot holding a radio at the start of a task looks identical to holding it at the end. Without memory of past actions or explicit stage tracking, the policy cannot distinguish these states and may execute incorrect actions.
    
    \item \textbf{No recovery demonstrations.} The training data consists of successful demonstrations only. When the robot deviates from demonstrated trajectories (inevitable given compounding errors), it encounters states never seen during training. The policy must somehow generalize to recover from these out-of-distribution situations.
    
    \item \textbf{Multi-modal action distributions.} Many states admit multiple valid action sequences (e.g., which hand to use, which object to grasp first). Different episodes of the same tasks were completed at different speeds in the training data.
\end{itemize}

\subsection{Our Approach}

We build upon Pi0.5~\cite{pi05_2025}, a vision-language-action (VLA) model that uses flow matching to predict action sequences. Our modifications address the challenges above through the following novel components:

\begin{itemize}
    \item \textbf{Modeling action structure.} Robot actions exhibit strong correlations---temporally (smooth trajectories) and across dimensions (coordinated joint movements). We model this structure explicitly by training with correlated noise sampled from $\mathcal{N}(0, \beta\boldsymbol{\Sigma} + (1-\beta)\mathbf{I})$, where $\boldsymbol{\Sigma}$ is the empirical action covariance and $\beta=0.5$. This makes training more efficient and enables principled inpainting during inference.
    
    \item \textbf{Providing non-Markovian context.} We introduce System 2 stage tracking: the model predicts the current task stage, and a voting mechanism filters noisy predictions to maintain stable stage estimates. This stage information is fused with task embeddings and fed back to the model, resolving ambiguous states.
    
    \item \textbf{Combining learning with heuristics.} Pure learning struggles with the lack of recovery data. We complement the learned policy with correction rules derived from failure analysis: simple heuristics that detect and recover from common failure modes like accidental gripper closures.
    
    \item  We apply \textbf{learnable mixed-layer attention} that allows each action expert layer to attend to learned linear combinations of all VLM layers rather than arbitrarily deciding how action expert layers should attend to VLM layers.
    
    \item  For training, we employ \textbf{multi-sample flow matching} (15 predictions per VLM forward pass) to reduce gradient variance while amortizing expensive vision-language computations.
    
    \item  At inference, we apply \textbf{action compression} via cubic splines to speed up action execution by 1.3$\times$.
    
    \item  We also simplified the VLM part by removing text processing and using \textbf{trainable task embeddings instead of the text prompt}. Technically this removes ``L'' from ``VLA'' and ``VLM'' terms but we keep the names ``VLA'' and ``VLM'' for simplicity.
\end{itemize}

\subsection{A Note on Evidence}

This report describes a competition entry, not a research paper. As a small independent team with limited compute, we prioritized winning over rigorous ablation studies. Many design choices were guided by intuition or quick experiments rather than systematic evaluation.

We present our methods honestly, without claiming that every component is necessary or optimal. That said, first place on the leaderboard suggests the overall approach is sound---even if we cannot isolate the contribution of each piece.

\subsection{Code and Models}

Our code for training and inference is available at \url{https://github.com/IliaLarchenko/behavior-1k-solution}.

We also share our model weights at \url{https://huggingface.co/IliaLarchenko/behavior_submission}.

This code and these weights should allow anyone to reproduce our results.

\section{Related Work}

Our work builds on vision-language-action models, flow matching, action chunking, and multi-task learning for robot control.

\subsection{Vision-Language-Action Models}

VLA models fine-tune pretrained vision-language models for robot control, leveraging semantic knowledge from web-scale pretraining. Early VLAs like RT-2~\cite{brohan2023rt2} use autoregressive token prediction for actions, which struggles with high-frequency control. Pi0~\cite{pi0_2024} (and later Pi0.5~\cite{pi05_2025}) uses flow matching for continuous action prediction, enabling dexterous manipulation at up to 50 Hz.

Using flow matching or diffusion as an action head is a popular choice in modern VLAs, but different architectures choose different ways to combine VLM and action head:

\begin{itemize}
    \item \textbf{Pi0.5}: Layer-to-layer expert-like attention. Layer $j$ of action expert attends to layer $j$ of VLM in a fused self-attention manner.
    \item \textbf{Gr00t}~\cite{gr00tn1_2025}: Attends only to the last VLM layer via standard cross-attention, like in the classic encoder-decoder transformer architecture~\cite{vaswani2017attention}.
    \item \textbf{SmolVLA}~\cite{shukor2025smolvla}: Keeps only half of the VLM layers, alternating self and cross attention layers.
\end{itemize}

We introduce learnable linear combinations of all VLM layers, letting the model decide optimal attention patterns rather than fixing them architecturally.

\subsection{Flow Matching and Diffusion for Actions}

Flow matching~\cite{lipman2023flow} and diffusion~\cite{chi2024diffusionpolicy} provide expressive action distributions that handle multi-modality better than regression. In Pi0.5, the flow-matching expert transforms normal noise into action chunks in multiple denoising steps.

Authors of Pi0 and Pi0.5 mention that first denoising steps are more difficult for the model to solve than the rest and introduced non-uniform sampling of flow-matching time during training to resolve this problem.

Our contribution in solving the same problem is modeling action correlations explicitly in the noise distribution. Standard flow matching uses $\boldsymbol{\epsilon} \sim \mathcal{N}(0, \mathbf{I})$; we use $\boldsymbol{\epsilon} \sim \mathcal{N}(0, \beta\boldsymbol{\Sigma} + (1-\beta)\mathbf{I})$ where $\boldsymbol{\Sigma}$ is the empirical action covariance and $\beta=0.5$. This makes training more efficient and enables correlation-guided inpainting.

\subsection{Action Chunking and Temporal Consistency}

ACT~\cite{zhao2023learning} introduced predicting action sequences (chunks) rather than single actions, improving temporal consistency and enabling open-loop execution. Real-time chunking methods~\cite{black2025rtc} use rolling windows with inpainting to balance consistency with reactivity---executing part of a predicted chunk while using the remainder as initial conditions for the next prediction.

Our correlation-aware inpainting uses learned action correlations to maintain smooth transitions between prediction windows, rather than treating inpainted and free actions independently.

\subsection{Multi-Task Learning for Robotics}

Training on multiple tasks can improve generalization and enable transfer learning. RT-1~\cite{brohan2022rt1} and RT-2~\cite{brohan2023rt2} and other VLA papers~\cite{pi0_2024,pi05_2025,gr00tn1_2025,shukor2025smolvla} demonstrate that large-scale multi-task training improves robustness and generalization in real-world robots.

Our multi-task training on 50 BEHAVIOR tasks shows similar benefits: the model learns recovery behaviors (e.g., grabbing fallen objects) that aren't explicitly present in single-task demonstrations. We hypothesize that training on diverse tasks exposes the model to a wide variety of states and transitions, enabling it to generalize recovery strategies across tasks.

\subsection{Hierarchical Reasoning for Long-Horizon Tasks}

Augmenting policies with high-level reasoning improves long-horizon performance. Many methods use separate models---a VLM for semantic subtask prediction and a low-level policy for execution. Hi Robot~\cite{shi2025hirobot} uses the same model for both high-level and low-level inference in a chain-of-thought style.

Our System 2 stage tracking is simpler: we predict task stages as an auxiliary output and use voting logic to filter noisy predictions. This provides non-Markovian context without explicit hierarchical planning or separate inference passes.

\subsection{Action Space Design}

\textbf{Delta actions} predict changes from current state rather than absolute positions, providing invariance to initial configuration, better generalization across starting states, and easier learning of smooth trajectories.

\textbf{Per-timestamp normalization}~\cite{tri2025lbm} addresses the fact that action distributions change over the prediction horizon---early actions in a chunk have small deltas while later actions are more varied. Normalizing per-timestamp makes the learning problem more uniform.

\subsection{BEHAVIOR Benchmark}

BEHAVIOR-1K~\cite{li2024behavior} is a benchmark featuring 1,000 everyday household activities in photo-realistic simulation (OmniGibson/Isaac Sim). The 2025 challenge includes 50 tasks with long-horizon execution (average 6.6 minutes, up to 14 minutes), complex object interactions, diverse activities (rearrangement, cooking, cleaning), and 10,000 expert demonstrations via teleoperation.

\section{Method Overview}

We organize our contributions into three main parts: \textbf{model architecture}, \textbf{training procedures}, and \textbf{inference optimizations}.

\subsection{High-Level Architecture}

\begin{figure*}[!htb]
  \centering
  \includegraphics[width=0.9\textwidth]{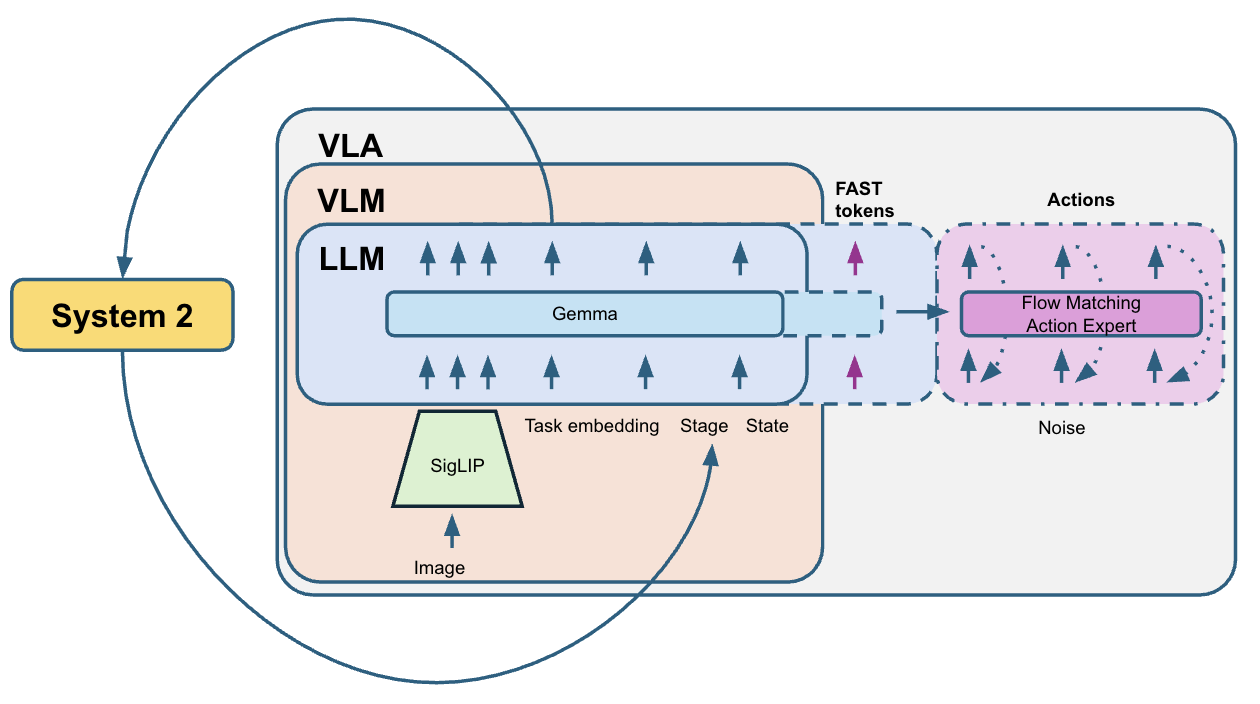}
  \caption{Overall architecture of our system. The vision backbone (SigLIP) processes images from three cameras, PaliGemma VLM processes visual tokens along with task embeddings and robot state, and the Action Expert predicts actions via flow matching. Task embeddings replace language processing, and System 2 provides stage information for non-Markovian context.}
  \label{fig:architecture}
\end{figure*}

Our model processes multi-camera RGB observations along with robot proprioceptive state to predict action sequences using flow matching (see Figure~\ref{fig:architecture}). The architecture consists of:

\begin{enumerate}
    \item \textbf{Vision Backbone}: SigLIP-So400m/14~\cite{zhai2023sigmoid} encoder processes images from three cameras (head, left wrist, right wrist)
    \item \textbf{Language Model Backbone}: PaliGemma-based~\cite{beyer2024paligemma} transformer processes visual tokens, task and stage information and robot state.
    \item \textbf{Action Expert}: Separate transformer that predicts action sequences via flow matching
\end{enumerate}

See Appendix~\ref{app:implementation} for more details.

\subsection{Our method description structure}

\paragraph{Part 1: Model Architecture (Section~\ref{sec:model}--\ref{sec:correlated_noise})}

We modify the Pi0.5 architecture in several ways:
\begin{enumerate}
    \item \textbf{Task Embeddings}: Replace language model processing with 50 trainable task embeddings (one per task)
    \item \textbf{KV Cache Transformation}: Learn linear combinations of VLM layer outputs for flexible mixed-layer attention
    \item \textbf{System 2 Stage Tracking}: Predict and track task stages to provide non-Markovian context
    \item \textbf{Custom Attention Masks}: Hierarchical attention pattern isolating reliable inputs from noisy ones
    \item \textbf{Correlated Noise Generation}: Sample noise from $\mathcal{N}(0, \beta\boldsymbol{\Sigma} + (1-\beta)\mathbf{I})$ instead of independent noise
\end{enumerate}

\paragraph{Part 2: Training Procedures (Section~\ref{sec:training})}

Our training innovations focus on reducing variance and improving action space learning:
\begin{enumerate}
    \item \textbf{Multi-Sample Flow Matching}: Compute 15 flow predictions per VLM forward pass with different noise/time samples
    \item \textbf{Delta Action Space}: Predict action deltas with per-timestamp normalization
    \item \textbf{Multi-Task Training}: Joint training on all 50 tasks with stage prediction as auxiliary task
    \item \textbf{Loss Composition}: Weighted combination of action loss, stage prediction loss, and FAST auxiliary loss
\end{enumerate}

\paragraph{Part 3: Inference Optimizations (Section~\ref{sec:inference})}

We introduce several inference-time techniques to improve action quality and efficiency:
\begin{enumerate}
    \item \textbf{Soft Correlation-Aware Inpainting}: Guide free dimensions during inpainting to maintain action correlation
    \item \textbf{Action Compression}: Interpolate 26 predicted actions to 20 execution steps using cubic splines (1.3$\times$ speedup)
    \item \textbf{Stage Voting}: Track stage transitions using majority voting over prediction history
    \item \textbf{Correction Rules}: Task-specific fixes for common failure modes (e.g., gripper recovery)
\end{enumerate}

\subsection{Mathematical Notation}

Throughout this report, we use the following notation:
\begin{itemize}
    \item $\mathbf{x}_t \in \mathbb{R}^{H \times D}$: Action sequence at flow time $t$, where $H=30$ is the action horizon and $D=23$ is the action dimension
    \item $\mathbf{a} \in \mathbb{R}^{H \times D}$: Target action sequence (ground truth)
    \item $\mathbf{v}_t \in \mathbb{R}^{H \times D}$: Predicted velocity (flow direction) at flow time $t$
    \item $\boldsymbol{\epsilon} \in \mathbb{R}^{H \times D}$: Noise vector
    \item $\boldsymbol{\Sigma} \in \mathbb{R}^{HD \times HD}$: Action covariance matrix (equivalent to correlation matrix since actions are normalized to zero mean and unit variance)
    \item $t \in [0, 1]$: Flow matching time parameter ($t=1$ is pure noise, $t=0$ is target)
    \item $\ell$: Timestep index in a trajectory (distinct from flow time $t$)
    \item $\tau \in \{0, 1, \ldots, 49\}$: Task ID
    \item $s \in \{0, 1, \ldots, s_{\max}(\tau)\}$: Task stage/substage
    \item $\mathbf{K}, \mathbf{V}$: Key and value caches from VLM layers
    \item $\mathcal{O}, \mathcal{U}$: Sets of inpainted (observed) and free (unobserved) action dimensions
\end{itemize}

\begin{figure*}[!htb]
  \centering
  \begin{subfigure}[b]{0.24\textwidth}
    \includegraphics[width=\textwidth]{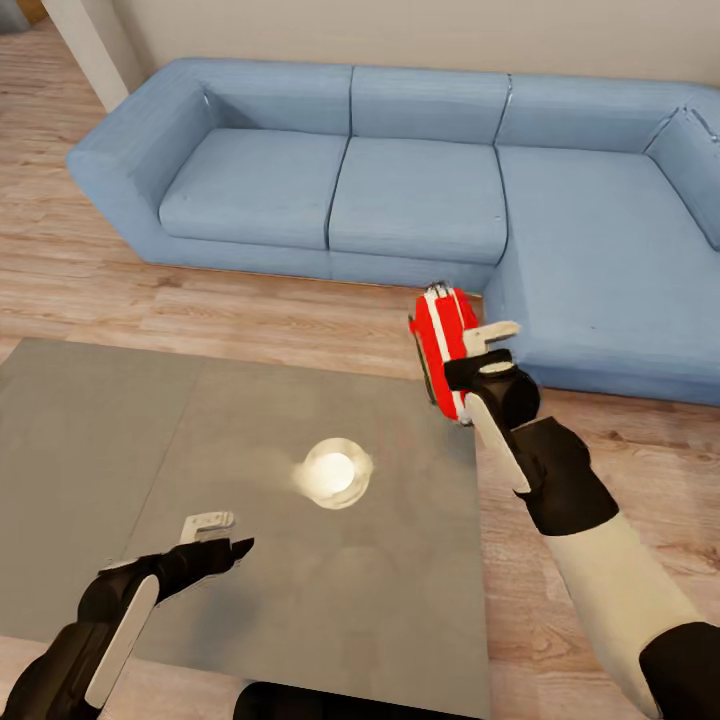}
    \caption{Lifting radio}
  \end{subfigure}
  \hfill
  \begin{subfigure}[b]{0.24\textwidth}
    \includegraphics[width=\textwidth]{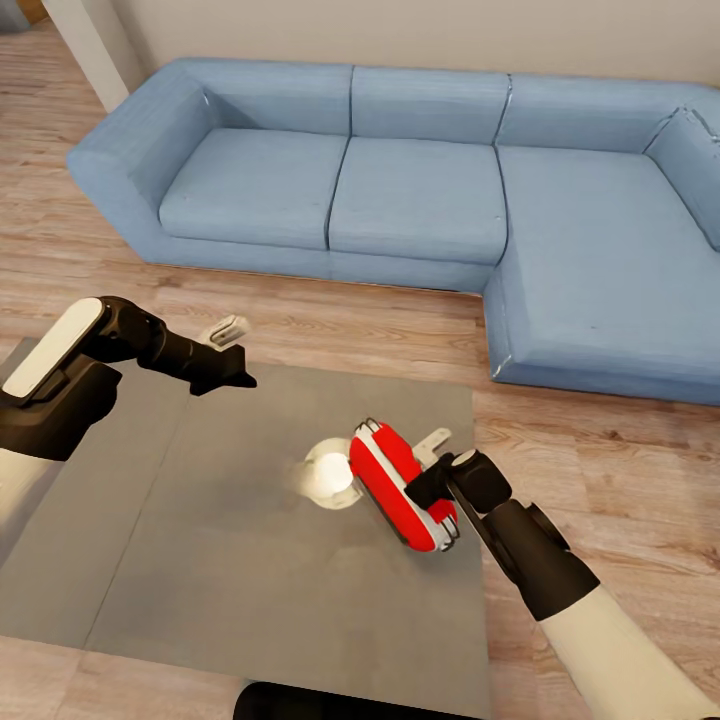}
    \caption{Placing radio back}
  \end{subfigure}
  \hfill
  \begin{subfigure}[b]{0.24\textwidth}
    \includegraphics[width=\textwidth]{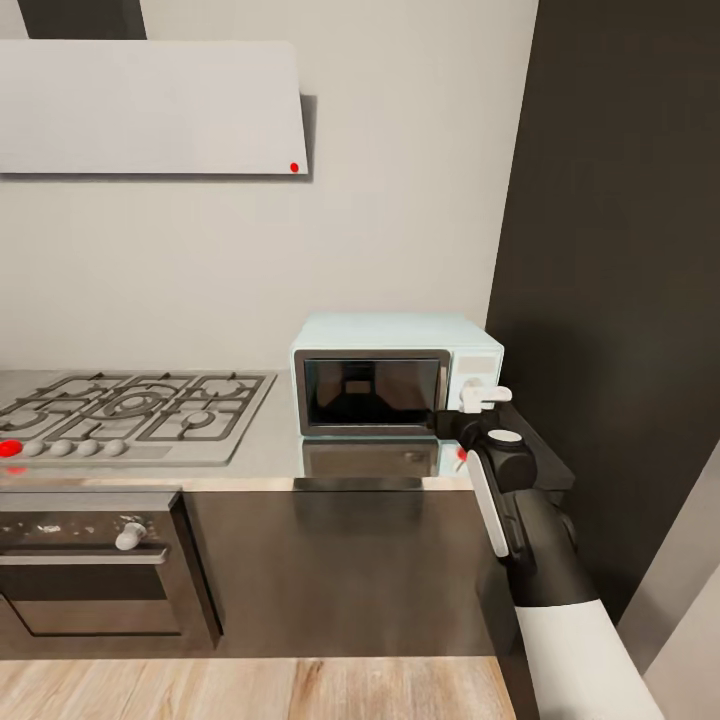}
    \caption{Opening microwave}
  \end{subfigure}
  \hfill
  \begin{subfigure}[b]{0.24\textwidth}
    \includegraphics[width=\textwidth]{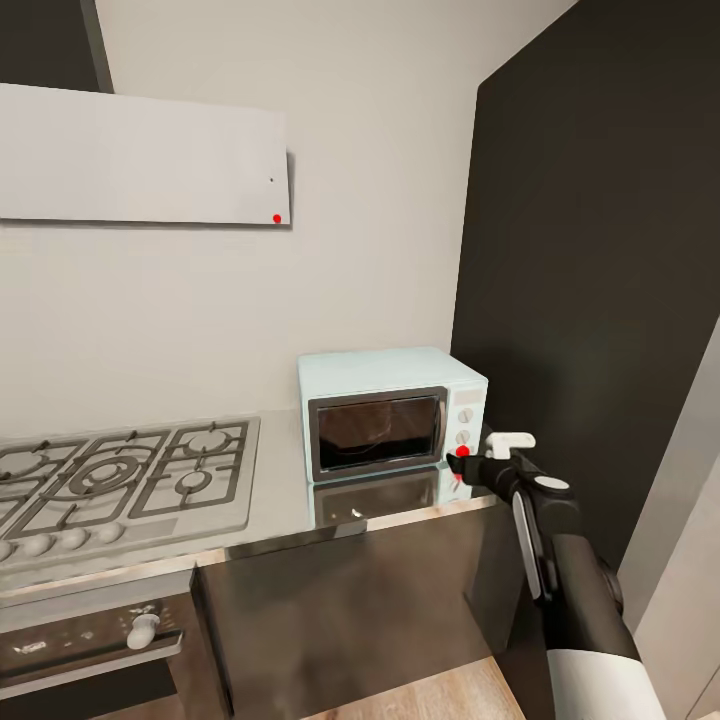}
    \caption{Pressing button}
  \end{subfigure}
  \caption{Examples of visually similar but semantically different states. (a,b) Radio task: lifting vs. placing back. (c,d) Popcorn task: opening microwave vs. pressing button after loading. Without stage tracking, the model confuses these states.}
  \label{fig:radio_popcorn}
\end{figure*}

\section{Model Architecture}
\label{sec:model}

This section describes the architectural modifications we made to the Pi0.5 base model to adapt it for BEHAVIOR-1K tasks.

\subsection{Task Embeddings}

The original Pi0.5 uses language embeddings for task specification. We replace this with task-specific embeddings for the structured BEHAVIOR-1K tasks.

The BEHAVIOR challenge requires very limited generalization. There are only 50 tasks that are present in both training and evaluation data. This means there is no explicit requirement for the policy to generalize to new tasks described in natural language.

Instead of processing natural language prompts, we use trainable task embeddings---one 2048-dimensional embedding for each of 50 tasks, trained from scratch.

This simplification is justified for BEHAVIOR-1K since:
\begin{enumerate}
    \item Only 50 distinct tasks (fixed set)
    \item Task semantics are implicit in demonstrations
    \item Removes language model processing overhead
    \item Allows the model to learn task-specific features directly
\end{enumerate}

\subsection{System 2: Stage Prediction and Fusion}

One of the big challenges that we faced across different problems is the existence of non-Markovian states. This means that the current state of the task is not enough to predict the correct next action. For example, the robot can see nearly identical images at the beginning and end of the same task (Figure~\ref{fig:radio_popcorn}).

This often confuses the model, which then acts incorrectly. To fix this issue we added a simple System 2 that predicts the current stage of the task based on the images and task embedding, applies voting logic to filter out incorrect predictions, and uses it as additional input to the model on following steps.

\subsubsection{Stage Prediction}

Each task is divided into 5--15 stages based on demonstration length. We predict the current stage using a linear classifier on the VLM output:
\begin{equation}
\mathbf{logits} = \mathbf{W}_{\text{stage}} \cdot \text{VLM}(\text{images, task\_id}) + \mathbf{b}_{\text{stage}}
\end{equation}
where $\mathbf{W}_{\text{stage}} \in \mathbb{R}^{15 \times 2048}$ (15 is the maximum number of stages across all tasks). Invalid stages for each task are masked with $-\infty$ before softmax.

Stage prediction achieves $\sim$99\% accuracy on training data, providing reliable context for action prediction.

\subsubsection{Stage-Task Fusion}

We fuse the task embedding with stage information using multiple learned representations (sine-cosine encoding, task-specific learned embeddings, and gated combinations). This provides 5 task-related tokens to the model. Details are in Appendix~\ref{app:stage_fusion}.

\begin{figure*}[!htb]
  \centering
  \begin{subfigure}[b]{0.48\textwidth}
    \includegraphics[width=\textwidth]{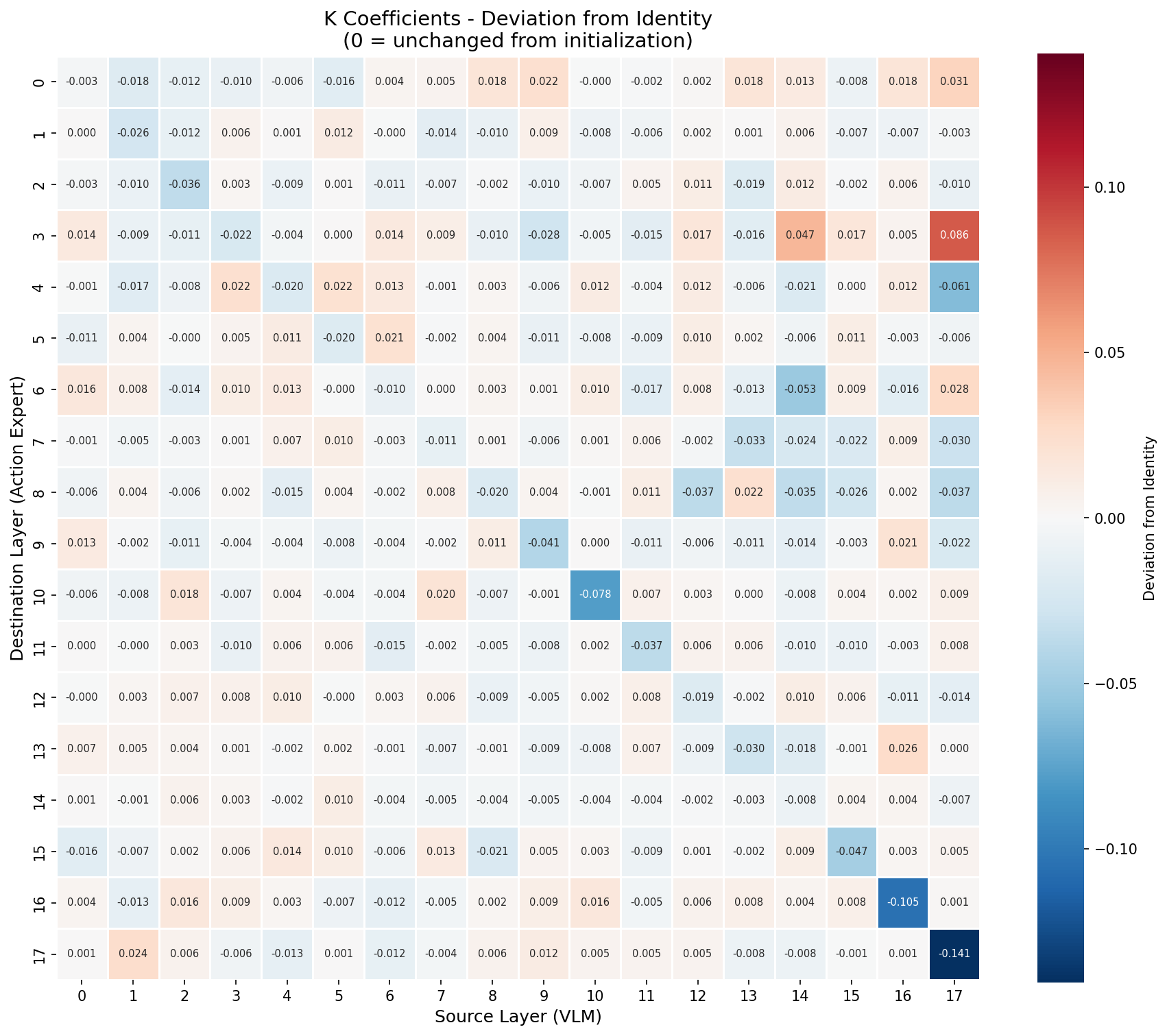}
    \caption{Key coefficients deviation from identity}
  \end{subfigure}
  \hfill
  \begin{subfigure}[b]{0.48\textwidth}
    \includegraphics[width=\textwidth]{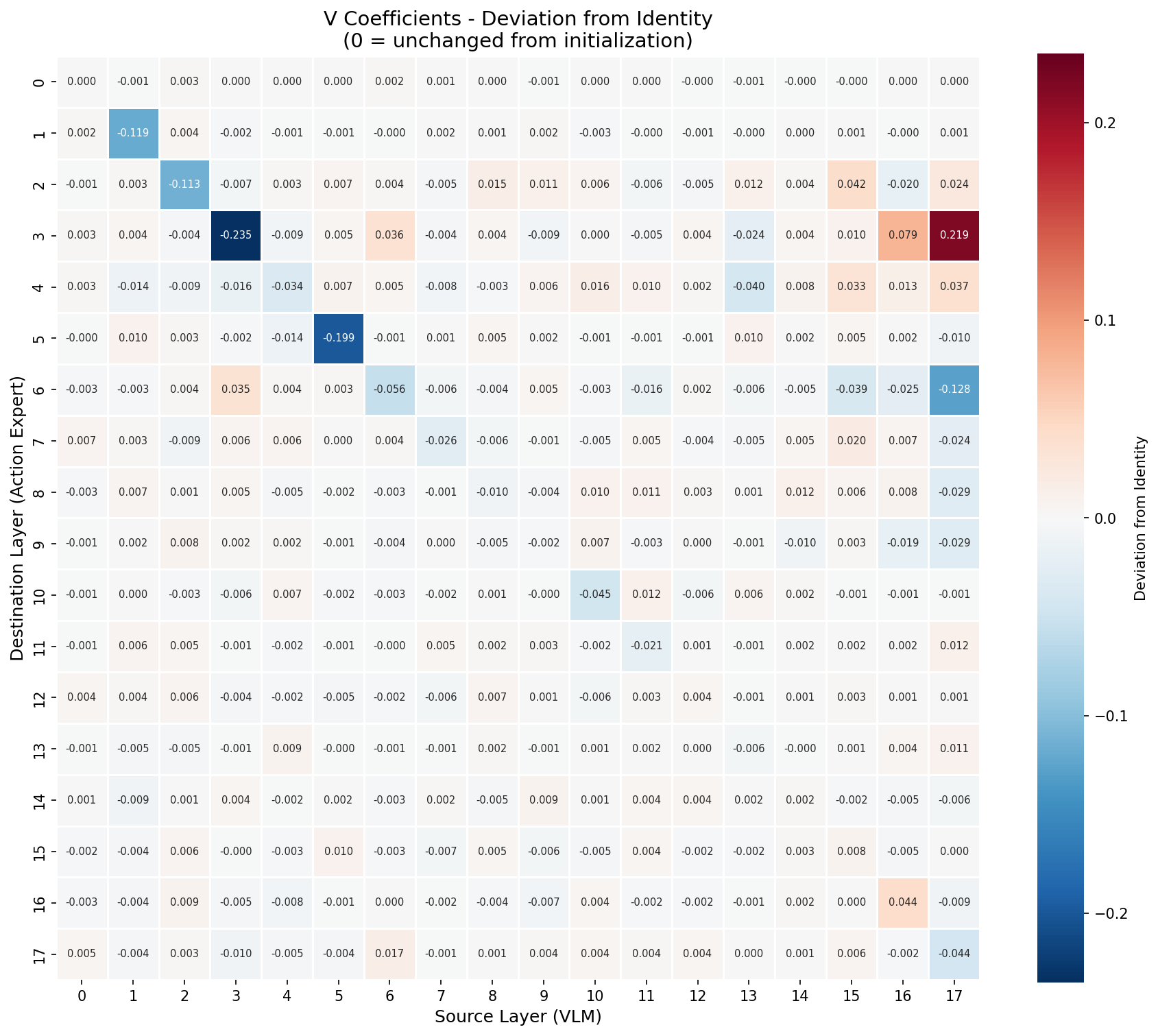}
    \caption{Value coefficients deviation from identity}
  \end{subfigure}
  \caption{Learned KV transformation weights showing deviation from initialization (identity). Each row corresponds to an action expert layer, each column to a VLM layer. Brighter colors indicate higher deviation.}
  \label{fig:kv_transform}
\end{figure*}

\subsection{KV Cache Transformation for Mixed-Layer Attention}

We don't see a clear winner among different ways how the flow-matching/diffusion action head attends to the VLM part in different VLA models, so we decided to let the model decide which layers to attend to and how.

\subsubsection{Learnable Layer Combination}

During both training and inference we first compute Key-Value cache for all layers of the VLM part. Then we transform it using learnable weights and biases.

For each action expert layer $j$, we compute transformed keys and values as linear combinations of all VLM layers:
\begin{align}
\mathbf{K}_j^{\text{new}} &= \sum_{i=1}^{L} w^{(K)}_{ij} \mathbf{K}_i + \mathbf{b}^{(K)}_j \\
\mathbf{V}_j^{\text{new}} &= \sum_{i=1}^{L} w^{(V)}_{ij} \mathbf{V}_i + \mathbf{b}^{(V)}_j
\end{align}
where $\mathbf{K}_i, \mathbf{V}_i \in \mathbb{R}^{B \times S \times N_h \times d_h}$ are key/value caches from VLM layer $i$ (with batch size $B$, sequence length $S$, $N_h$ attention heads, head dimension $d_h$), $w^{(K)}_{ij}, w^{(V)}_{ij} \in \mathbb{R}$ are learnable scalar weights, $\mathbf{b}^{(K)}_j, \mathbf{b}^{(V)}_j \in \mathbb{R}^{N_h \times d_h}$ are learnable bias terms, and $L$ is the number of VLM layers.

\textbf{Initialization}: $w_{ij} = \delta_{ij}$ (identity), $\mathbf{b}_j = \mathbf{0}$, so the model starts with Pi0.5's layer-to-layer attention.

\subsubsection{Properties}

This design allows the model to: (1) attend to any VLM layer---weights $w_{ij}$ can select early, middle, or late layers; (2) form smooth combinations by attending to weighted averages of multiple layers; (3) learn from data without manual architecture search. The approach is parameter-efficient: for each of 18 action expert layers, we learn only 18 scalar coefficients plus one bias tensor $\in \mathbb{R}^{N_h \times d_h}$, separately for keys and values.

We use different coefficients for keys and values, as there is no reason for them to be identical. Figure~\ref{fig:kv_transform} shows the learned deviations from identity initialization. Since we initialized from pretrained Pi0.5 weights after significant fine-tuning, the identity initialization was already a good prior. We observe some tendency to attend more to the last VLM layer, though this could be noise. We expect this approach to have larger effects for models trained from scratch or initialized from non-robotics VLMs.

\subsection{Custom Attention Masks}

\begin{figure}[!htb]
  \centering
  \includegraphics[width=0.8\linewidth]{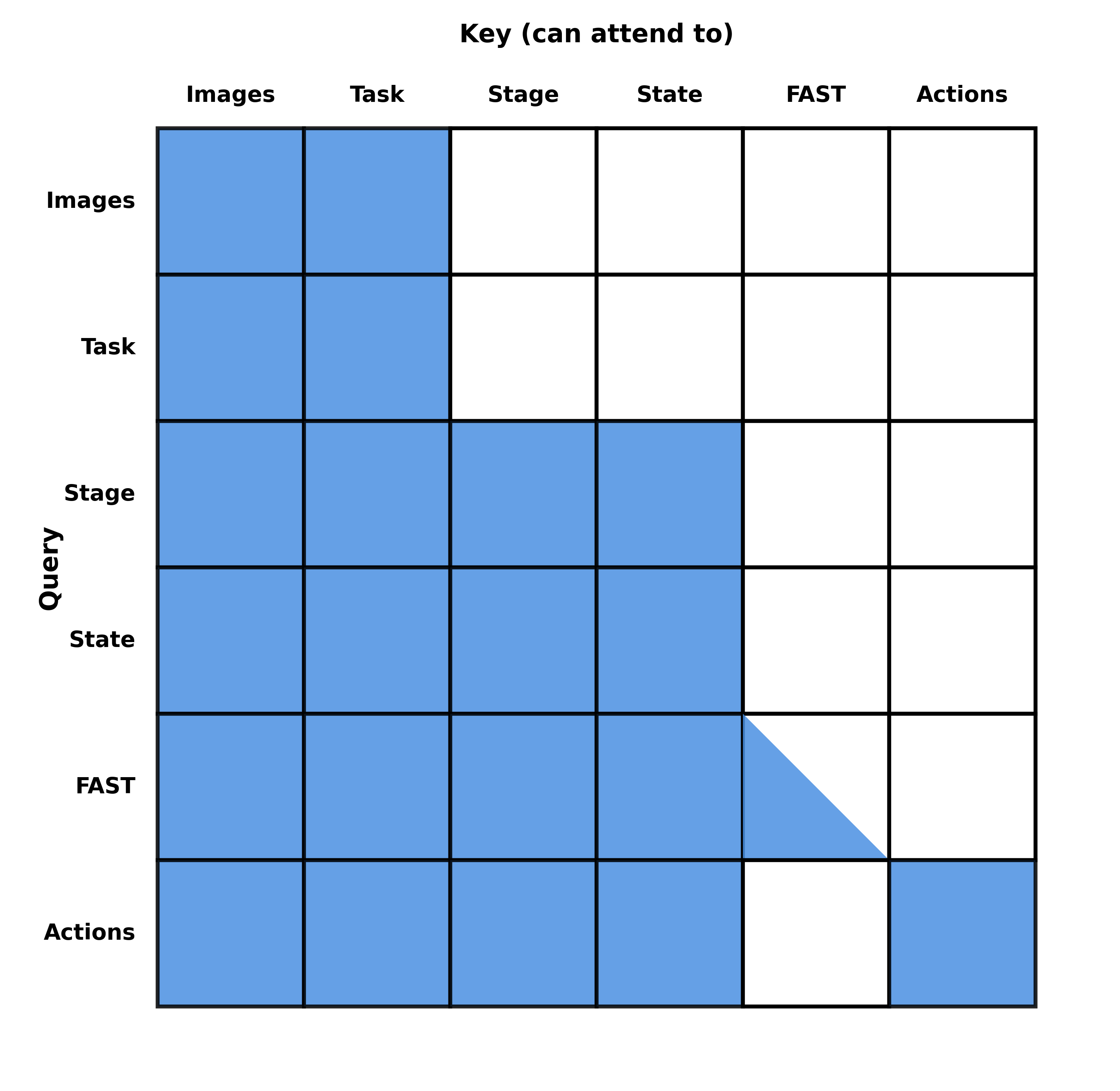}
  \caption{Attention mask structure during training. Blue indicates bidirectional attention, white indicates no attention, and diagonal patterns indicate causal attention.}
  \label{fig:attention_mask}
\end{figure}

We use a hierarchical attention pattern that isolates reliable inputs from noisy ones:

\begin{enumerate}
    \item \textbf{Image tokens}: Bidirectional attention among themselves and task token
    \item \textbf{Task token}: Bidirectional with images
    \item \textbf{Stage tokens}: Attend to images, task, and proprio state
    \item \textbf{State tokens}: Attend to images, task, stage, and other state tokens
    \item \textbf{FAST tokens} (if used): Attend to all prefix tokens and causally to each other
    \item \textbf{Action expert tokens}: Bidirectional among themselves, attend to all other prefix tokens (except FAST)
\end{enumerate}

\textbf{Rationale}: 
\begin{enumerate}
    \item Images and task embeddings are the most reliable inputs (deterministic from observations). We prevent them from attending to noisier inputs like robot state (which can drift during inference) or predicted stage (which could be incorrect). This keeps visual features clean. 
    \item Only images and task embeddings are used in System 2 to predict the current stage; to avoid target leakage, they should not attend to stage tokens. 
    \item FAST tokens predict actions autoregressively, so they attend to all prefix tokens and causally to each other (used only during training). 
    \item Action expert tokens predict the whole chunk simultaneously, so they use bidirectional attention among themselves while attending to all prefix tokens except FAST. See Figure~\ref{fig:attention_mask}.
\end{enumerate}

\subsection{Delta Action Space with Per-Timestamp Normalization}

\subsubsection{Delta Actions}

Instead of predicting absolute joint positions, we predict \textbf{delta actions}:
\begin{equation}
\mathbf{a}_{\text{delta}}[\ell+i] = \mathbf{a}_{\text{abs}}[\ell+i] - \mathbf{q}_{\text{current}}[\ell]
\end{equation}
where $\mathbf{q}$ represents joint positions, $i \in \{0, \ldots, H-1\}$ is the index within the action chunk, and $\ell$ is the current timestep.

\subsubsection{Per-Timestamp Normalization}

For each action dimension $d$ and index in the chunk $i$:
\begin{equation}
\mu_{di} = \text{mean}(\mathbf{a}_{\text{delta}}[i, d]), \quad \sigma_{di} = \text{std}(\mathbf{a}_{\text{delta}}[i, d])
\end{equation}
Normalized actions:
\begin{equation}
\tilde{\mathbf{a}}[i, d] = \frac{\mathbf{a}_{\text{delta}}[i, d] - \mu_{di}}{\sigma_{di}}
\end{equation}

\textbf{Why per-timestamp?} Action distributions change over time within a trajectory: initial actions in the chunk are very close to the current state (small delta), while later actions are more varied. Per-timestamp normalization makes the learning problem more uniform across the horizon.

\textbf{Note:} velocities and gripper positions are excluded from per-timestamp normalization.

\subsection{Action Expert Architecture}

The action expert uses the Gemma 300M architecture ($\sim$300M parameters, 18 layers), identical to Pi0.5. It attends to the transformed VLM KV cache and outputs velocity predictions $\mathbf{v}_t = \text{ActionExpert}(\mathbf{x}_t, t, \text{KV}_{\text{VLM}})$. The training loss is $\mathcal{L}_{\text{action}} = \mathbb{E}_{t, \boldsymbol{\epsilon}} [ \|\mathbf{v}_t - (\boldsymbol{\epsilon} - \mathbf{a})\|^2 ]$. See Appendix~\ref{app:action_expert} for architecture details.

\subsection{FAST Auxiliary Training}

We implemented FAST~\cite{pertsch2025fast} auxiliary training with loss weight 0.05. Since we removed text input, we directly train FAST token embeddings mapping to PaliGemma input. FAST uses global quantile normalization for better DCT compression because per-timestamp normalization breaks temporal smoothness of actions. See Appendix~\ref{app:fast} for details.

\section{Correlated Noise for Flow Matching}
\label{sec:correlated_noise}

One of our key innovations is modeling action correlations explicitly during flow matching training and inference.

\subsection{Motivation: Action Correlations in Robot Control}

\begin{figure}[!htb]
  \centering
  \includegraphics[width=0.77\linewidth,keepaspectratio=false,height=0.7\linewidth]{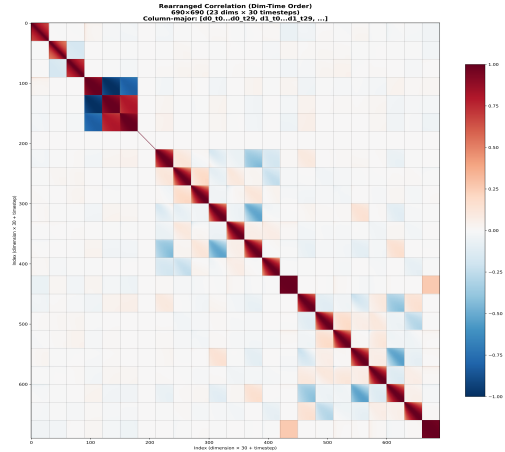}
  \caption{Action correlation matrix $\boldsymbol{\Sigma} \in \mathbb{R}^{690 \times 690}$ computed from training data (30 timesteps $\times$ 23 action dimensions = 690). The matrix shows strong block-diagonal structure indicating high temporal correlation and significant cross-dimensional correlation.}
  \label{fig:correlation_matrix}
\end{figure}

Robot actions exhibit strong correlations in two ways (see Figure~\ref{fig:correlation_matrix}):
\begin{enumerate}
    \item \textbf{Temporal correlation}: Actions at nearby timesteps are similar (smooth trajectories)
    \item \textbf{Cross-dimensional correlation}: Joint velocities are coordinated (e.g., trunk joints move together)
\end{enumerate}

Standard flow matching uses independent Gaussian noise $\boldsymbol{\epsilon} \sim \mathcal{N}(0, \mathbf{I})$. This creates a training problem: the early denoising steps ($t \approx 1$) are difficult, but once a few denoising steps are made and the model learns correlation structure, later prediction becomes trivial.

\textbf{Our solution}: Generate correlated noise $\boldsymbol{\epsilon} \sim \mathcal{N}(0, \boldsymbol{\Sigma})$ that already matches the action structure. This makes all denoising steps more balanced in difficulty.

\subsection{Correlation Matrix Estimation}

We estimate the full correlation matrix from the training set. Let $\mathbf{a}^{(n)} \in \mathbb{R}^{H \times D}$ be the $n$-th normalized action sequence (flattened to $\mathbb{R}^{HD}$). The sample covariance is:
\begin{equation}
\hat{\boldsymbol{\Sigma}} = \frac{1}{N} \sum_{n=1}^{N} \text{vec}(\mathbf{a}^{(n)}) \cdot \text{vec}(\mathbf{a}^{(n)})^T
\end{equation}
where $\text{vec}(\cdot)$ flattens the $H \times D$ matrix to a $HD$-dimensional vector.

\subsection{Beta Shrinkage for Robustness}

Using pure $\boldsymbol{\Sigma}$ can be unstable. We apply \textbf{shrinkage regularization}:
\begin{equation}
\boldsymbol{\Sigma}_{\text{reg}} = \beta \boldsymbol{\Sigma} + (1 - \beta) \mathbf{I}
\end{equation}
where $\beta \in [0, 1]$ is the shrinkage parameter. We use $\beta = 0.5$ as a balanced choice.

\subsection{Correlated Noise Generation}

To sample $\boldsymbol{\epsilon} \sim \mathcal{N}(0, \boldsymbol{\Sigma}_{\text{reg}})$, we use the Cholesky decomposition:
\begin{equation}
\boldsymbol{\Sigma}_{\text{reg}} = \mathbf{L} \mathbf{L}^T
\end{equation}
where $\mathbf{L}$ is lower triangular. Then:
\begin{equation}
\boldsymbol{\epsilon} = \mathbf{L} \mathbf{z}, \quad \mathbf{z} \sim \mathcal{N}(0, \mathbf{I})
\end{equation}

\subsection{Impact on Training}

With correlated noise, the flow matching interpolation becomes:
\begin{equation}
\mathbf{x}_t = t \cdot \boldsymbol{\epsilon} + (1-t) \cdot \mathbf{a}, \quad \boldsymbol{\epsilon} \sim \mathcal{N}(0, \boldsymbol{\Sigma}_{\text{reg}})
\end{equation}

\textbf{At $t=1$ (pure noise)}: $\mathbf{x}_1 = \boldsymbol{\epsilon}$ has the same correlation structure as real actions. The model sees plausible action patterns even at the noisiest step.

\textbf{At $t \in (0, 1)$}: The interpolation maintains correlation structure throughout the denoising process.

This makes the training task more uniform and keeps difficulties of different denoising steps more balanced.

\section{Training}
\label{sec:training}

\subsection{Training Data}

We train on the BEHAVIOR-1K demonstration dataset: 10,000 expert demonstrations across 50 tasks (1200+ hours total). Each demonstration is divided into 5--15 stages based on temporal position for stage prediction training.

\subsection{Multi-Sample Flow Matching}

Standard flow matching computes one action prediction per observation, sampling $(t, \boldsymbol{\epsilon})$ randomly for each batch element. This introduces significant variance in the training signal.

\subsubsection{Motivation}

The flow matching loss has two sources of randomness:
\begin{enumerate}
    \item \textbf{Time sampling}: $t \sim \text{Beta}(1.5, 1)$
    \item \textbf{Noise sampling}: $\boldsymbol{\epsilon} \sim \mathcal{N}(0, \boldsymbol{\Sigma}_{\text{reg}})$
\end{enumerate}

We can amortize the expensive VLM forward pass across multiple flow samples and reduce the randomness of the resulting gradient.

\subsubsection{Algorithm}

For each training batch:
\begin{enumerate}
    \item \textbf{VLM forward pass} (once): Compute KV cache for all prefix tokens
    \item \textbf{Multi-sample action prediction} ($N=15$ times): Sample different $(t_n, \boldsymbol{\epsilon}_n)$ for each sample, compute noisy actions, run action expert
    \item \textbf{Backward pass}: Gradients flow back through all $N$ samples
\end{enumerate}

\subsection{Loss Function}

Our total loss is a weighted combination of three components:
\begin{equation}
\mathcal{L}_{\text{total}} = \mathcal{L}_{\text{action}} + \lambda_s \mathcal{L}_{\text{stage}} + \lambda_f \mathcal{L}_{\text{FAST}}
\end{equation}

\subsubsection{Action Loss}

The flow matching loss averaged over $N$ samples:
\begin{equation}
\mathcal{L}_{\text{action}} = \frac{1}{N} \sum_{n=1}^{N} \frac{1}{HD} \sum_{h=1}^{H} \sum_{d=1}^{D} \left( v_{t_n}[h,d] - u_{t_n}[h,d] \right)^2
\end{equation}

\subsubsection{Stage Prediction Loss}

Cross-entropy loss for stage classification with weight $\lambda_s = 0.1$.

\subsubsection{FAST Auxiliary Loss}

Weight $\lambda_f = 0.05$ (reduced from 0.1 after initial training).

\subsection{Multi-Task vs. Task-Specific Training}

We train models in two steps:

\textbf{1. Multi-task (initial)}: Train on all 50 tasks simultaneously. This phase took 15 days of non-stop training on 8$\times$H200 GPUs.

\textbf{2. Task-group-specific (fine-tuning)}: We split tasks based on validation results: best (highest success rate), good (score $>$ 0), bad (score $\sim$0) and trained them separately. This phase took $\sim$1 week per group.

The final submission uses 4 task-specific checkpoints, automatically switching based on task ID.

\subsection{Training Budget}

Our total competition budget was $\sim$\$13k. We spent $\sim$\$3k of personal money on experiments and evaluation. \$10k was sponsored by Nebius and we used it for the main training runs on 8$\times$H200 GPUs.

\section{Inference Optimizations}
\label{sec:inference}

\subsection{Correlation-Aware Inpainting}

To ensure smooth action sequences and resolve local multi-modality, we use a rolling inpainting strategy.

\begin{figure*}[!htb]
  \centering
  \includegraphics[width=0.85\textwidth]{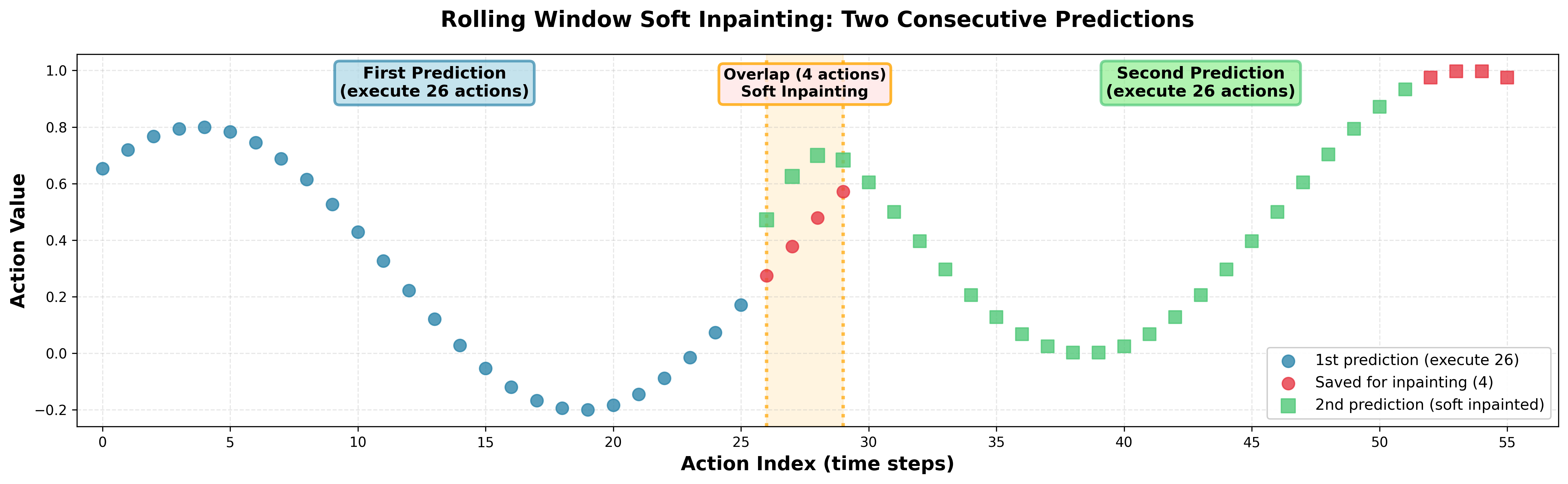}
  \caption{Rolling window soft inpainting across two consecutive predictions. The first prediction generates 30 actions (blue circles), executes 26, and saves the last 4 (red circles) for inpainting. The second prediction (green squares) starts close to the saved actions through soft inpainting.}
  \label{fig:inpainting}
\end{figure*}

\begin{figure*}[!htb]
  \centering
  \includegraphics[width=0.85\textwidth]{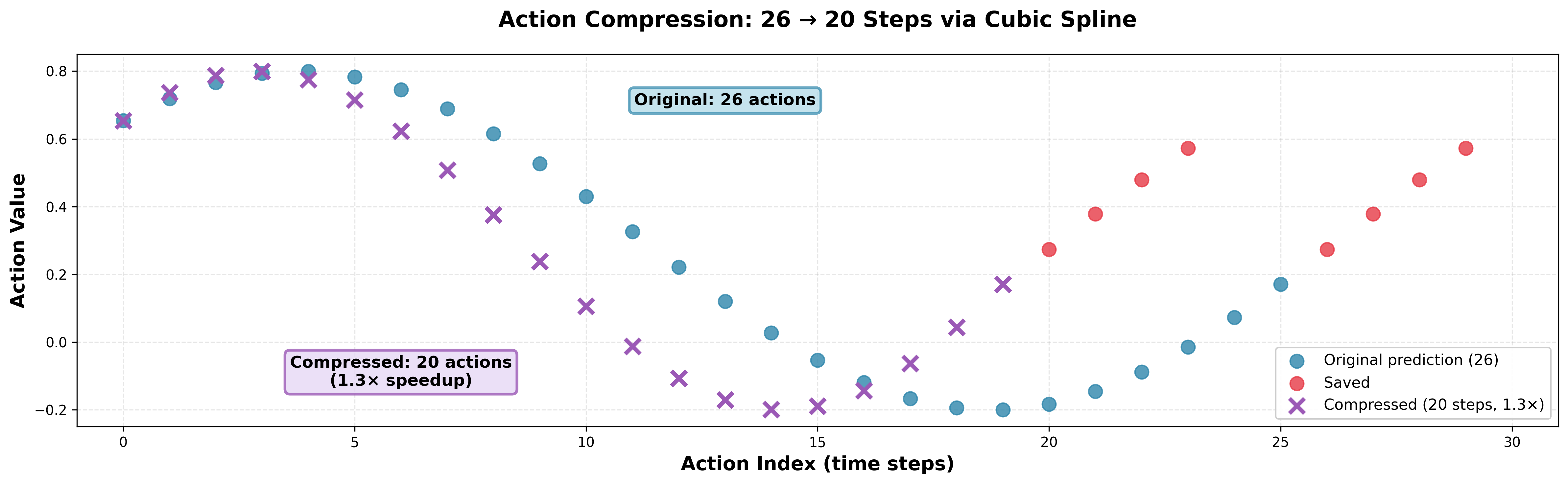}
  \caption{Action compression via cubic spline interpolation. The original 26 actions (blue circles) are compressed to 20 steps (purple crosses), achieving 1.3$\times$ speedup.}
  \label{fig:compression}
\end{figure*}

\subsubsection{Soft Inpainting Strategy}

Instead of executing all predicted actions directly we:
\begin{enumerate}
    \item \textbf{Predict} 30 actions: $\{\mathbf{a}_0, \mathbf{a}_1, \ldots, \mathbf{a}_{29}\}$
    \item \textbf{Execute} first 26 actions: $\{\mathbf{a}_0, \ldots, \mathbf{a}_{25}\}$
    \item \textbf{Save} last 4 actions: $\{\mathbf{a}_{26}, \ldots, \mathbf{a}_{29}\}$ as initial conditions
    \item \textbf{Next prediction}: Generate 30 new actions such that first 4 almost match the saved actions
    \item \textbf{Repeat}
\end{enumerate}

\subsubsection{Correlation-Aware Correction Propagation}

The key challenge is: how to constrain the initial actions while respecting the correlation structure? A naive approach applies hard constraints on the first 4 actions with no adjustment for the rest. This creates discontinuity at the boundary between actions 4 and 5---the model predictions don't respect correlation between inpainted and free actions, and the input at every flow-matching step becomes out-of-distribution.

Our approach propagates corrections using the learned correlation structure. Let $\mathcal{O}$ = indices of inpainted actions (first 4 timesteps, $|\mathcal{O}| = 4 \times 23 = 92$ dimensions), $\mathcal{U}$ = indices of free actions ($|\mathcal{U}| = 26 \times 23 = 598$ dimensions). Partition the correlation matrix:
\begin{equation}
\boldsymbol{\Sigma}_{\text{reg}} = \begin{bmatrix} \boldsymbol{\Sigma}_{\mathcal{OO}} & \boldsymbol{\Sigma}_{\mathcal{OU}} \\ \boldsymbol{\Sigma}_{\mathcal{UO}} & \boldsymbol{\Sigma}_{\mathcal{UU}} \end{bmatrix}
\end{equation}

At each denoising step with time $t$, after the model predicts $\mathbf{x}_t^{\text{pred}}$:
\begin{enumerate}
    \item Compute desired state on inpainted dimensions using saved initial actions $\mathbf{x}_0^{\mathcal{O}}$ and fixed noise $\mathbf{z}^{\mathcal{O}}$: $\mathbf{x}_t^{\text{desired}, \mathcal{O}} = (1-t) \mathbf{x}_0^{\mathcal{O}} + t \mathbf{z}^{\mathcal{O}}$
    \item Compute additive correction: $\boldsymbol{\delta}^{\mathcal{O}} = \mathbf{x}_t^{\text{desired}, \mathcal{O}} - \mathbf{x}_t^{\text{pred}, \mathcal{O}}$
    \item Apply hard constraint on inpainted dimensions
    \item Propagate correction to free dimensions: $\boldsymbol{\delta}^{\mathcal{U}} = \mathbf{M}_{\text{corr}} \boldsymbol{\delta}^{\mathcal{O}}$ where $\mathbf{M}_{\text{corr}} = \boldsymbol{\Sigma}_{\mathcal{UO}} \boldsymbol{\Sigma}_{\mathcal{OO}}^{-1}$
    \item Apply the correlated correction: $\mathbf{x}_t^{\mathcal{U}} = \mathbf{x}_t^{\text{pred}, \mathcal{U}} + \boldsymbol{\delta}^{\mathcal{U}}$
\end{enumerate}

The correction matrix $\mathbf{M}_{\text{corr}}$ encodes how much each free dimension should adjust given a correction on inpainted dimensions, ensuring smooth transitions. We precompute this matrix once using a numerically stable solver. See Figure~\ref{fig:inpainting}.

\subsubsection{Soft Inpainting via Time Threshold}

We apply inpainting correction only during early denoising steps ($t > 0.3$). At early $t$, maintaining constraints and correlation is crucial for smooth trajectories. At late $t$ (close to the target), the model should have full freedom to adapt to current observations. This ``soft'' inpainting allows deviation from the initial plan if observations change.

\subsection{Action Compression via Cubic Spline Interpolation}

Speeding up action execution relative to demonstration data can improve task completion rates by allowing more prediction cycles per episode and more attempts to recover from failures. This idea appeared in prior work~\cite{larchenko2025dot} (2$\times$ speedup) and Figure AI's ``sport mode''~\cite{figure2025helix} (linear resampling of action chunks). We apply a similar principle with cubic spline interpolation (see Figure~\ref{fig:compression}):

\begin{itemize}
    \item \textbf{Predicted}: 26 actions at 30Hz
    \item \textbf{Executed}: 20 steps at 30Hz
    \item \textbf{Speedup}: $26/20 = 1.3\times$
\end{itemize}

We use cubic spline interpolation to generate smooth intermediate actions rather than linear resampling, which can introduce jitter.

\textbf{Velocity scaling}: We scale base velocity dimensions by $1.3\times$ to account for faster execution: $\mathbf{a}_{\text{scaled}}[0:3] = \mathbf{a}_{\text{interpolated}}[0:3] \times 1.3$. Joint velocities (arms, trunk) remain unchanged since they are already normalized and the controller handles timing.

\textbf{Adaptive compression}: We disable compression when gripper states change significantly. Many failures are related to grasping, so we slow down and give the policy more time when the robot tries to grasp an object.

\subsection{Stage Tracking with Voting Logic}

\begin{figure*}[!htb]
  \centering
  \includegraphics[width=0.85\textwidth]{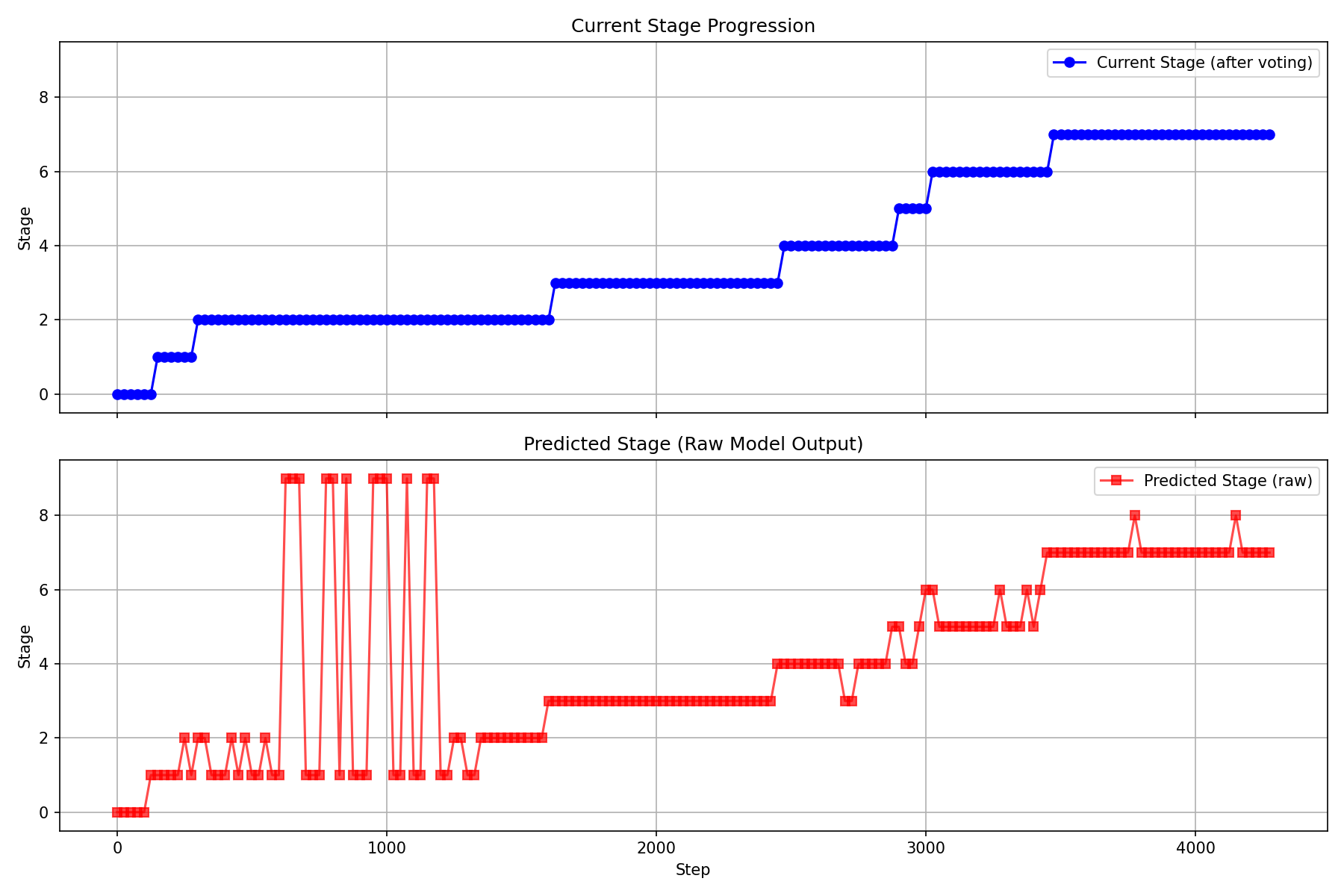}
  \caption{Stage tracking with voting logic resolves non-Markovian states. \textbf{Top}: Current stage after voting shows smooth, stable progression. \textbf{Bottom}: Raw model predictions are noisy and sometimes jump ahead (e.g., predicting stage 9 when still in early stages). In the radio task example, the robot grasping the radio at the beginning (stage 1-2) looks visually identical to placing it back later (stage 9), but System 2 correctly maintains stage 2 context despite model confusion.}
  \label{fig:stage_jump}
\end{figure*}

The model predicts the current stage at each inference step. Since individual predictions can be noisy (see Figure~\ref{fig:stage_jump}), we employ a majority voting scheme to ensure stable stage transitions.

We maintain a sliding window of the three most recent stage predictions. At each inference step, the model outputs stage logits, and we take the argmax to obtain the predicted stage $\hat{s}$. This prediction is appended to the history buffer. Stage transitions follow three rules based on the prediction history:

\textbf{Forward transition}: If at least two of three predictions indicate the next stage $s_{\text{curr}} + 1$, we advance to that stage and clear history. This majority voting prevents premature transitions from single noisy predictions while still allowing responsive progression.

\textbf{Skip detection}: If all three predictions consistently indicate stage $s_{\text{curr}} + 2$, this suggests the robot has completed a stage faster than expected or the stage was already satisfied. We advance by one stage to catch up, then clear history.

\textbf{Rollback}: If all three predictions unanimously indicate the previous stage $s_{\text{curr}} - 1$, we roll back by one stage. This handles cases where a subtask needs to be re-attempted. Requiring unanimous agreement makes rollback more conservative than forward transitions.

After any stage transition, the prediction history is cleared to prevent stale predictions from influencing future transitions.

\subsection{Correction Rules}

\subsubsection{General Gripper Correction}

BEHAVIOR-1K dataset is very clean and doesn't contain recovery demonstrations. In practice, if the policy fails any action, there is a very high chance of ending up in an out-of-distribution state and getting completely stuck.

One of the most common failures across all tasks is to fail a grasp and close the gripper on empty air. There are almost no training data where the robot opens the gripper after closing it---this leads to complete failure as the robot is stuck and can't perform any actions.

\textbf{Note}: After training on 50 tasks, we observed that recovery behavior actually emerged in some cases---the robot learned to open the gripper even without explicit training data for it. However, these cases were rare and didn't improve the total success rate substantially.

To resolve this problem, we implemented a simple rule: if the gripper is closed and it was never closed in training data for this particular task at the same stage, we treat it as a failed grasp and completely open the gripper.

\textbf{Impact}: This correction rule alone approximately \textbf{doubled our success rate} in selected tasks where grabbing objects was a common failure mode.

\subsubsection{Task-Specific Rules}

We started experimenting with more fine-grained task-specific rules but eventually added only one simple rule for the ``turning\_on\_radio'' task (roll back by 2 stages if we reach the final stage but didn't succeed). We believe this approach has potential for improving results but is not scalable or generalizable, so we didn't pursue it further.

\section{Evaluation}
\label{sec:evaluation}

\subsection{Evaluation Setup}

\subsubsection{Challenge Protocol}

The BEHAVIOR-1K challenge uses a standardized evaluation protocol:
\begin{itemize}
    \item \textbf{Tasks}: 50 household activities (same as in the demo dataset)
    \item \textbf{Episodes per task}: 10 evaluation episodes (fixed instances with randomized initial conditions)
    \item \textbf{Success metric}: Goal condition satisfaction (binary and partial)
    \item \textbf{Time limit}: Task-specific ($2 \times$ average human task completion time in the demo dataset)
\end{itemize}

\subsubsection{Evaluation Instances}

Each task has 10 pre-defined evaluation instances with randomized object positions (task-relevant objects) and robot initial pose.

Public leaderboard scores use 10 published instances; private scores are computed on an additional held-out set of 10 instances per task known only to the organizers.

Evaluation instances are distinct from training demonstrations (different seeds but drawn from the same distribution), testing generalization to new spatial configurations. The required sequence of actions stays mostly the same as in the demo dataset; there is no test of unseen object categories, language goal understanding, or entirely new tasks.

\textbf{Note}: During train data collection there was an unintentional bias toward simpler instances. The public and private evaluation instances do not share this bias and are on average more challenging.

\subsection{Metrics}

\subsubsection{Primary Metric: Averaged Partial Success}

Each task has a list of goal conditions. Partial success is the fraction of those conditions satisfied at the end of an episode---how far the robot progressed toward full completion.
\begin{itemize}
    \item \textbf{Episode-level partial success}: fraction of goal conditions satisfied in that episode
    \item \textbf{Task-level score}: average episode partial success across the 10 evaluation episodes for that task
    \item \textbf{Overall score}: average task-level partial success across the 50 tasks (q\_score)
\end{itemize}

Binary success (all goal conditions satisfied) is measured and reported for analysis, but it is \textbf{not} the ranking metric.

\subsubsection{Secondary Metrics: Efficiency}

The challenge also tracks efficiency metrics (not used for ranking):
\begin{itemize}
    \item \textbf{Simulated time}: Total simulation steps $\times$ time per step
    \item \textbf{Distance navigated}: Cumulative base movement distance
    \item \textbf{Hand displacement}: Cumulative gripper movement distance
\end{itemize}

\subsection{Evaluation Infrastructure}

Running all 500 episodes (tasks $\times$ instances) on a single machine would take $\sim$10 days, so we parallelized.

Parallelization options considered:
\begin{itemize}
    \item Multiple environments in one process (experimental in OmniGibson; avoided to keep the official eval script unchanged)
    \item Many simulator instances on one machine (impractical due to high RAM use and a memory leak)
    \item Multiple machines (chosen): elastic access to inexpensive, RTX-capable GPUs with stable setup; we used RunPod and scaled to 20 RTX 4090s in parallel
\end{itemize}

As an independent team without existing infrastructure, we built a lightweight infrastructure that fit our budget and kept iteration speed high.

Resulting evaluation infrastructure:
\begin{itemize}
    \item On-demand cluster to avoid paying for idle compute; full evaluation finishes in under 2 days
    \item Heavy assets (simulation code, model checkpoints) on a shared network volume. Using a default Docker container allows us to spin up new machines in minutes
    \item Evaluation outputs (videos, metrics, logs) automatically synced to an S3 bucket (we chose Cloudflare R2) with a simple UI for browsing models' evaluation results
    \item A lightweight offline balancer spreads tasks evenly and reruns episodes that crash (e.g., simulator segfaults)
\end{itemize}

\section{Results}
\label{sec:results}

\subsection{Competition Results}

Top 5 teams evaluated on held-out instances (see Table~\ref{tab:results}).

\begin{table*}[!htb]
\centering
\small
\begin{tabular}{clccccc}
\toprule
Rank & Team & Affiliation & \multicolumn{2}{c}{Binary Success} & \multicolumn{2}{c}{q\_score} \\
\cmidrule(lr){4-5} \cmidrule(lr){6-7}
 & & & public & private & public & private \\
\midrule
1 & \textbf{Robot Learning Collective (ours)} & Independent & 0.1120 & \textbf{0.1240} & 0.2605 & \textbf{0.2599} \\
2 & Comet & NVIDIA Research & 0.1440 & 0.1140 & 0.1830 & 0.2514 \\
3 & SimpleAI Robot & Beijing Simple AI Technology Co Ltd & 0.1400 & 0.1080 & 0.1943 & 0.1591 \\
4 & The North Star & Huawei CRI EAI Team & 0.1280 & 0.0760 & 0.1702 & 0.1204 \\
5 & Embodied Intelligence & Independent & 0.0620 & 0.0520 & 0.1110 & 0.0947 \\
\bottomrule
\end{tabular}
\caption{2025 BEHAVIOR Challenge  leaderboard results. Full leaderboard available at \url{https://behavior.stanford.edu/challenge/leaderboard.html}.}
\label{tab:results}
\end{table*}

\begin{figure}[!htb]
  \centering
  \includegraphics[width=\linewidth]{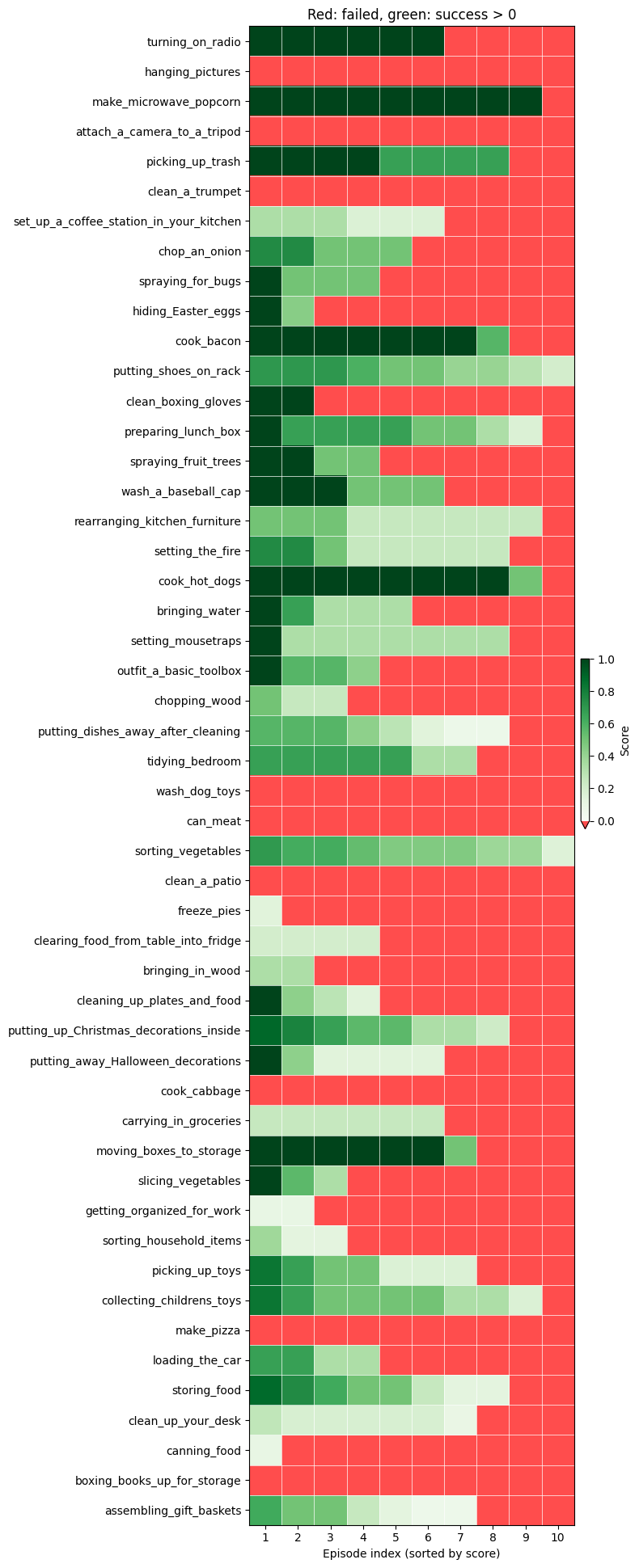}
  \caption{Per-episode scores on public evaluation. Each row is a task (sorted by average duration in demo dataset), each column an episode (sorted by score). Dark green indicates success, light green - partial success, red indicates failure.}
  \label{fig:per_task_scores}
\end{figure}

We achieved a 26\% q\_score with negligible difference between public and private evaluation. In our case, partial successes contribute roughly half of the total score (see Figure~\ref{fig:per_task_scores}).

\subsection{Analysis}

\subsubsection{General Observations}
\begin{itemize}
    \item Some tasks are almost solved, except under particularly tricky initial conditions
    \item For tasks with 0 success, we do not observe that they are generally impossible; instead, they usually contain one tricky step that involves very high-precision manipulation (with low success rate even for human teleoperators) or a carefully followed sequence that is slightly beyond the current model's limits
    \item Task duration does not appear to be a fundamental obstacle: longer tasks simply have many more steps, which makes full success harder, but partial success remains very achievable
\end{itemize}

\subsubsection{Failure Mode Analysis}

To understand what goes wrong when the robot fails, we labeled a subset of tasks (15/50) with multiple-choice failure reasons (see Figure~\ref{fig:failure_reasons}):

\begin{figure}[!htb]
  \centering
  \includegraphics[width=\linewidth]{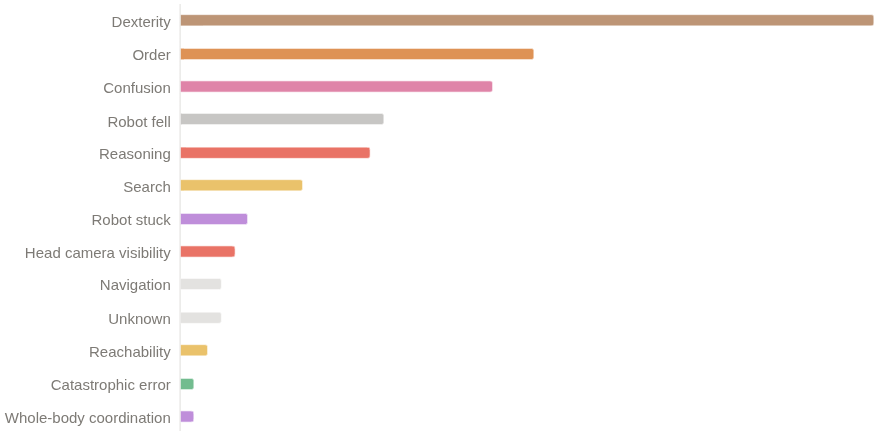}
  \caption{Distribution of failure reasons across labeled evaluation episodes. Dexterity issues (clumsiness in grasping/releasing) dominate, followed by progress/order errors and confusion from out-of-distribution states.}
  \label{fig:failure_reasons}
\end{figure}

\begin{itemize}
    \item \textbf{Dexterity}: Clumsiness---the robot cannot reliably pick up or release items. This is the dominant problem, accounting for around one-third of failures
    \item \textbf{Order}: Progress errors---wrong action sequence. Many tasks require a particular order. Other common issue is deciding to finish early, exacerbated by ``tails'' on passive actions in the demo dataset
    \item \textbf{Confusion}: Weird behavior probably caused by the robot entering an out-of-distribution state
    \item \textbf{Robot fell}: While attempting to squat to pick an item from the floor, the robot sometimes starts to fall backward
    \item \textbf{Reasoning}: The robot should choose a locally non-obvious action (e.g., go around the table instead of trying to reach through it)
    \item \textbf{Search}: Randomness in the denoising process works surprisingly well for continued exploration, but can lead to repeatedly traversing the same area
    \item \textbf{Robot stuck}: The model sometimes keeps trying to move while the robot base contact prevents it
    \item \textbf{Head camera visibility}: The head camera does not cover the floor in front of the robot or shelves above the head
    \item \textbf{Navigation}: Navigating to fixed locations (same as in the demo dataset) is generally not a problem, but in rare cases the model confuses the task and tries to go to the location of another task.
    \item \textbf{Unknown}: Failures that we could not attribute to any specific problem.
    \item \textbf{Reachability}: The robot is realistically limited in reach; sometimes—especially after unsuccessful grasps—this makes task completion impossible.
    \item \textbf{Catastrophic error}: The robot makes a mess of the environment, rendering the task unachievable.
    \item \textbf{Whole-body coordination}: The robot lacks awareness that an arm may be stuck, which prevents further safe navigation.
\end{itemize}

Many problems are amplified by the specifics of the data‑collection procedure: operating a humanoid‑like robot in simulation is challenging for human operators. Tasks were divided into steps; if the operator failed at a given step, the simulation was rolled back to the start of that substep. This enabled collection of complex, long‑horizon demonstrations, but the resulting trajectories are very idealized and cover only a narrow manifold—small deviations from the ideal trajectory often cause the model to output random actions or freeze.

\subsubsection{Additional Observations}

\textbf{Emerging recovery behavior from cross-task learning}: This was a key factor in improving the model. Single-task models show no recovery behavior. The same architecture trained on all 50 tasks exhibits a wide range of recovery behaviors, such as picking up fallen objects.

In general, more training improves success rates across tasks, but for certain tasks the multi-task model performs worse. We hypothesize that this may be due to short task duration and thus low relative weight in the dataset, or because of the confusion between different tasks with similar visual features.

During the main training run we periodically branched off checkpoints and fine‑tuned them on 1–2 tasks. Early in training this gave a significant performance boost, but at later stages the main run reached comparable performance, suggesting that undertraining was the main limiting factor.

\subsubsection{Summary}
Our analysis highlights problems that are currently a major focus for practitioners building real‑world systems: VLA models for dexterous manipulation, System‑2‑style components to guide IL policies, and diverse pre‑training datasets to widen the manifold where the model produces meaningful actions. This suggests that work on this challenge is relevant to real‑world problems.

\subsection{Ablations}

We did not have the budget to run full ablations, but our small-scale experiments show:

\begin{itemize}
    \item The model is surprisingly tolerant to image quality. Comparing 224$\times$224 generation vs. 720$\times$720 downscaled did not lead to meaningful changes
    \item More surprisingly, machines in our cloud provider used for evaluation did not support NGX (due to Docker settings), which led to easily noticeable image‑quality degradation, but this had very small impact on success rate. As the private score was calculated with the correct setting, we decided to do nothing about it.
    \item Advantage‑weighting dataset rows based on subtask duration (following the intuition that if a step took a long time for the operator, demonstration quality tends to be lower) did not show a success‑rate improvement on the one task we tried (although we noticed a significant difference in task duration, likely corresponding to different operators)
    \item Small variations in inference parameters (number of actions to execute, execution speedup, voting history length) did not show significant change. Extreme changes (like execution of only 10 steps for 30 steps chunk, or speed up 2 times) led to the score degradation
    \item Gripper opening correction rule showed 2.2$\times$ increase in q-score for a subset of 13 tasks with 3 episodes each (39 episodes total)
\end{itemize}

\section{Discussion and Conclusion}
\label{sec:discussion}

We demonstrated how to adapt Pi0.5 for the BEHAVIOR Challenge, achieving 1st place with 26\% q-score across 50 diverse household tasks. Our approach combines architectural innovations (correlated noise, learnable mixed-layer attention, System 2 stage tracking) with practical optimizations (action compression, correction rules).

\textbf{The result is far from optimal.} None of our models fully converged---despite training nearly continuously for a month, only $\sim$2 epochs passed through the dataset. Longer training would almost certainly improve results. The challenge timeline, not model capacity, was the limiting factor.

\textbf{Task-specific checkpoints helped but are they necessary?} We used 4 checkpoints primarily to optimize our remaining time before the deadline (even though we have seen positive effect from the task confusion reduction). We believe comparable results are achievable with a single multi-task model, which would be more elegant and practical for deployment. Comparing single model vs. multiple task-specific fine-tuned models is an interesting research question.

\textbf{Gripper correction rules were surprisingly effective.} This simple heuristic---detecting and recovering from accidental gripper closures---contributed significantly to our final score. More task-specific rules could likely boost performance further, though such rules are competition-specific and not generalizable research contributions.

\textbf{We propose several novel ideas but lack proper ablations.} Due to resource constraints, we could not isolate the contribution of each component. Rigorous ablation studies would be valuable to identify which innovations actually matter and which are redundant.

\textbf{Pretrained VLA models are essential.} Our attempts to train a smaller model (same architecture but smaller transformers) from scratch didn't show promising results. Leveraging Pi0.5's pretrained weights was critical---the VLA paradigm of ``pretrain on internet scale, fine-tune for robotics'' remains the most practical path forward.

Long-horizon household manipulation remains challenging. Even with extensive engineering, we achieve only 26\% success.

\subsection{Future Directions}

The following directions seem promising to us:
\begin{itemize}
    \item \textbf{Collecting corrective actions}: We believe that even a comparatively small number of hours of demonstrations collected from states where the model fails can significantly improve its robustness
    \item \textbf{Offline RL with advantage reweighting}: This seems a promising approach to train the model to avoid unfavorable states
    \item \textbf{VLM as System 2}: Modern VLMs exhibit good video understanding. With no strict budget on using cloud APIs, powerful VLMs prompted with task-specific ``checklists'' may offer a strong effort/results trade-off
    \item \textbf{RL in simulation using privileged information with teacher--student distillation}: This should work, but is not directly applicable to real-world problems
    \item \textbf{Robot control constraints to prevent falling}: Implementing a safety envelope so that actions produced by the model do not cause the robot to fall
    \item \textbf{Fish-eye head camera}: As well as an additional camera on the back of the robot should help
\end{itemize}

\section*{Acknowledgments}

We would like to thank following people for their help and support: Vladimir Ershov, Justyna Ilczuk, Andrey Mulenkov.

Special thanks to \textbf{Nebius} (\url{https://nebius.com/}) for providing cloud GPU compute resources and sponsoring the development of our solution.

\bibliographystyle{plainnat}

\section*{Appendix}
\appendix

\section{Stage-Task Fusion Details}
\label{app:stage_fusion}

We fuse the task embedding with stage information using multiple learned representations. We don't advocate for this specific approach and didn't conduct proper ablation studies due to time and resource constraints---we simply added different representations hoping some would work well.

\textbf{Input representations}:
\begin{enumerate}
    \item \textbf{Task embedding}: Base task vector $\mathbf{e}_{\tau} \in \mathbb{R}^{2048}$
    \item \textbf{Sine-cosine stage encoding}: $\mathbf{s}_{\text{sincos}} \in \mathbb{R}^{1024}$ (positional encoding of normalized stage)
    \item \textbf{Learned stage embedding}: $\mathbf{s}_{\text{learned}} \in \mathbb{R}^{1024}$ (task-specific trainable embeddings)
\end{enumerate}

\textbf{Stage normalization} (task-specific):
\begin{equation}
s_{\text{norm}} = \frac{s}{\max(s_{\max}(\tau) - 1, 1)}
\end{equation}
where $s_{\max}(\tau) \in [5, 15]$ is the number of stages for task $\tau$. This ensures stage 0 $\rightarrow$ 0.0 and final stage $\rightarrow$ 1.0 for all tasks.

\textbf{Gated fusion mechanism}: We concatenate all inputs and learn sigmoid gates to modulate each component:
\begin{equation}
\mathbf{x}_{\text{all}} = [\mathbf{e}_{\tau} \,;\, \mathbf{s}_{\text{sincos}} \,;\, \mathbf{s}_{\text{learned}}] \in \mathbb{R}^{4096}
\end{equation}

\textbf{Four fusion strategies}:
\begin{enumerate}
    \item \textbf{Task-gated}: $\mathbf{f}_1 = \mathbf{e}_{\tau} \odot \mathbf{g}_{\text{task}}$
    \item \textbf{Balanced fusion}: $\mathbf{f}_2 = \mathbf{W}_5 \cdot \text{ReLU}(\mathbf{W}_4 \mathbf{x}_{\text{all}} + \mathbf{b}_4) + \mathbf{b}_5$
    \item \textbf{Stage-dominant}: $\mathbf{f}_3 = \mathbf{W}_6 \cdot [(\mathbf{s}_{\text{sincos}} \odot \mathbf{g}_{\text{sincos}}) \,;\, (\mathbf{s}_{\text{learned}} \odot \mathbf{g}_{\text{learned}})] + \mathbf{b}_6$
    \item \textbf{Pure stage}: $\mathbf{f}_4 = [\mathbf{s}_{\text{sincos}} \,;\, \mathbf{s}_{\text{learned}}]$
\end{enumerate}

All four representations have dimension 2048. The final output is $[\mathbf{f}_1 \,;\, \mathbf{f}_2 \,;\, \mathbf{f}_3 \,;\, \mathbf{f}_4] \in \mathbb{R}^{4 \times 2048}$. These 4 stage-conditioned tokens are passed to the model along with the base task embedding (total: 5 task-related tokens).

\section{Implementation Details}
\label{app:implementation}

\begin{itemize}
    \item \textbf{Action Space}: 23D continuous actions (3D base velocity, 4D trunk, 7D left arm, 1D left gripper, 7D right arm, 1D right gripper)
    \item \textbf{Action Horizon}: 30 timesteps
    \item \textbf{Image Resolution}: 224$\times$224 RGB
    \item \textbf{Vision Backbone}: Frozen SigLIP-So400m/14
    \item \textbf{Framework}: JAX
    \item \textbf{Training}: FSDP model parallelism on 8$\times$H200 GPUs
    \item \textbf{Inference}: Single GPU (tested on RTX 4090, 24GB VRAM)
\end{itemize}

\section{Action Expert Architecture}
\label{app:action_expert}

The action expert uses the Gemma 300M architecture (identical to Pi0.5):
\begin{itemize}
    \item \textbf{Layers}: 18
    \item \textbf{Hidden dimension}: 1024
    \item \textbf{Feedforward dimension}: 4096
    \item \textbf{Attention heads}: 8 (with 1 KV head using grouped-query attention)
    \item \textbf{Head dimension}: 256
    \item \textbf{Parameters}: $\sim$311M (action expert only)
\end{itemize}

Input sequence consists of noisy actions $\mathbf{x}_t = t \cdot \boldsymbol{\epsilon} + (1-t) \cdot \mathbf{a}$ projected to 1024D, plus time embedding $\text{posemb\_sincos}(t, 1024)$ processed by 2-layer MLP (AdaRMS conditioning).

\section{FAST Auxiliary Training}
\label{app:fast}

Since we removed text input and the Gemma tokenizer (replaced with task embeddings), we directly train FAST token embeddings rather than reusing the language model's vocabulary:

\begin{itemize}
    \item \textbf{FAST token embedding layer}: Maps discrete FAST tokens (vocab size 1024) to PaliGemma width (2048D)
    \item \textbf{FAST projection head}: Projects VLM outputs back to FAST vocabulary for next-token prediction
    \item \textbf{Action encoding}: We encode 22 action dimensions (dimensions 0:6, 7:23) using a trained FAST tokenizer, excluding the padded dimensions and one dimension with 0 variance
    \item \textbf{Normalization}: FAST uses global quantile normalization to preserve temporal smoothness for better DCT compression, even when the rest of the model uses per-timestamp normalization
    \item \textbf{Sequence processing}: FAST tokens are processed autoregressively with causal attention, attending to all prefix tokens (images, task embeddings, stages, state)
\end{itemize}

The auxiliary loss is computed as standard cross-entropy on the predicted token sequence. The final model uses a FAST loss weight of \textbf{0.05}.

\textbf{Note}: Overall, we believe that FAST may be more beneficial for VLA pre-training and have a limited impact on fine-tuning. But we speculate (without proper ablations) that it helps with training the model to complete simpler tasks like navigation faster and prevents the model from ``forgetting'' them while the action expert focuses on dexterous manipulation tasks.

\end{document}